\documentclass[11pt,letterpaper]{article}
\usepackage[letterpaper]{geometry}
\usepackage{acl2012}
\usepackage{times}
\usepackage{latexsym}
\usepackage{amsmath}
\usepackage{multirow}
\usepackage{url}
\makeatletter
\newcommand{\@BIBLABEL}{\@emptybiblabel}
\newcommand{\@emptybiblabel}[1]{}
\makeatother
\usepackage[hidelinks]{hyperref}
\DeclareMathOperator*{\argmax}{arg\,max}
\usepackage{graphicx}
\usepackage{subfigure}
\usepackage{relsize}
\usepackage{xspace}
\usepackage{booktabs}
\usepackage[textsize=small]{todonotes}

\newcommand{\SwissProt}{\textsc{Swiss-Prot}\xspace}
\newcommand{\BiDAF}{\emph{BiDAF}\xspace}
\newcommand{\DrugBank}{\textsc{DrugBank}\xspace}
\newcommand{\FastQA}{\emph{FastQA}\xspace}
\newcommand{\MEDLINE}{\textsc{Medline}\xspace}
\newcommand{\MedHop}{\textsc{MedHop}\xspace}
\newcommand{\TriviaQA}{\emph{TriviaQA}\xspace}
\newcommand{\WikiHop}{\textsc{WikiHop}\xspace}

\newcommand{\WikiReadingUnseen}{54.4\%\xspace}
\newcommand{\WikiReading}{\textsc{WikiReading}\xspace}
\newcommand{\Wikidata}{\textsc{Wikidata}\xspace}
\newcommand{\Wikipedia}{\textsc{Wikipedia}\xspace}

\newif\ifarxiv
\arxivfalse

\title{
    Constructing Datasets \\
    for Multi-hop Reading Comprehension Across Documents
}

\author{
    {Johannes Welbl$^1$ \hspace{5em} Pontus Stenetorp$^1$ \hspace{5em} Sebastian Riedel$^{1,2}$ \vspace{0.5em}}\\
    {\centering $^1$University College London,~ $^2$Bloomsbury AI} \\
    {\centering \tt \{j.welbl,p.stenetorp,s.riedel\}@cs.ucl.ac.uk}
}

\date{}
\begin{document}
\maketitle

\begin{abstract}
    Most Reading Comprehension methods limit themselves to queries which can be answered using a single sentence, paragraph, or document.
    Enabling models to combine disjoint pieces of textual evidence would extend the scope of machine comprehension methods, but currently no resources exist to train and test this capability.
    We propose a novel task to encourage the development of models for text understanding across multiple documents and to investigate the limits of existing methods.
    In our task, a model learns to seek and combine evidence -- effectively performing multi-hop, alias multi-step, inference.
    We devise a methodology to produce datasets for this task, given a collection of query-answer pairs and thematically linked documents.
    Two datasets from different domains are induced,\footnote{Available at \url{http://qangaroo.cs.ucl.ac.uk}} 
    and we identify potential pitfalls and devise circumvention strategies.
    We evaluate two previously proposed competitive models and find that one can integrate information across documents.
    However, both models struggle to select relevant information; and providing documents guaranteed to be relevant greatly improves their performance.
    While the models outperform several strong baselines, their best accuracy reaches 
    54.5\% on an annotated test set, compared to human performance at 
    85.0\%, leaving ample room for improvement.
\end{abstract}

\section{Introduction}
\begin{figure}[t]
    \centering
    \includegraphics[width=1.0\columnwidth,trim={0 8.5cm 0 0cm},clip]{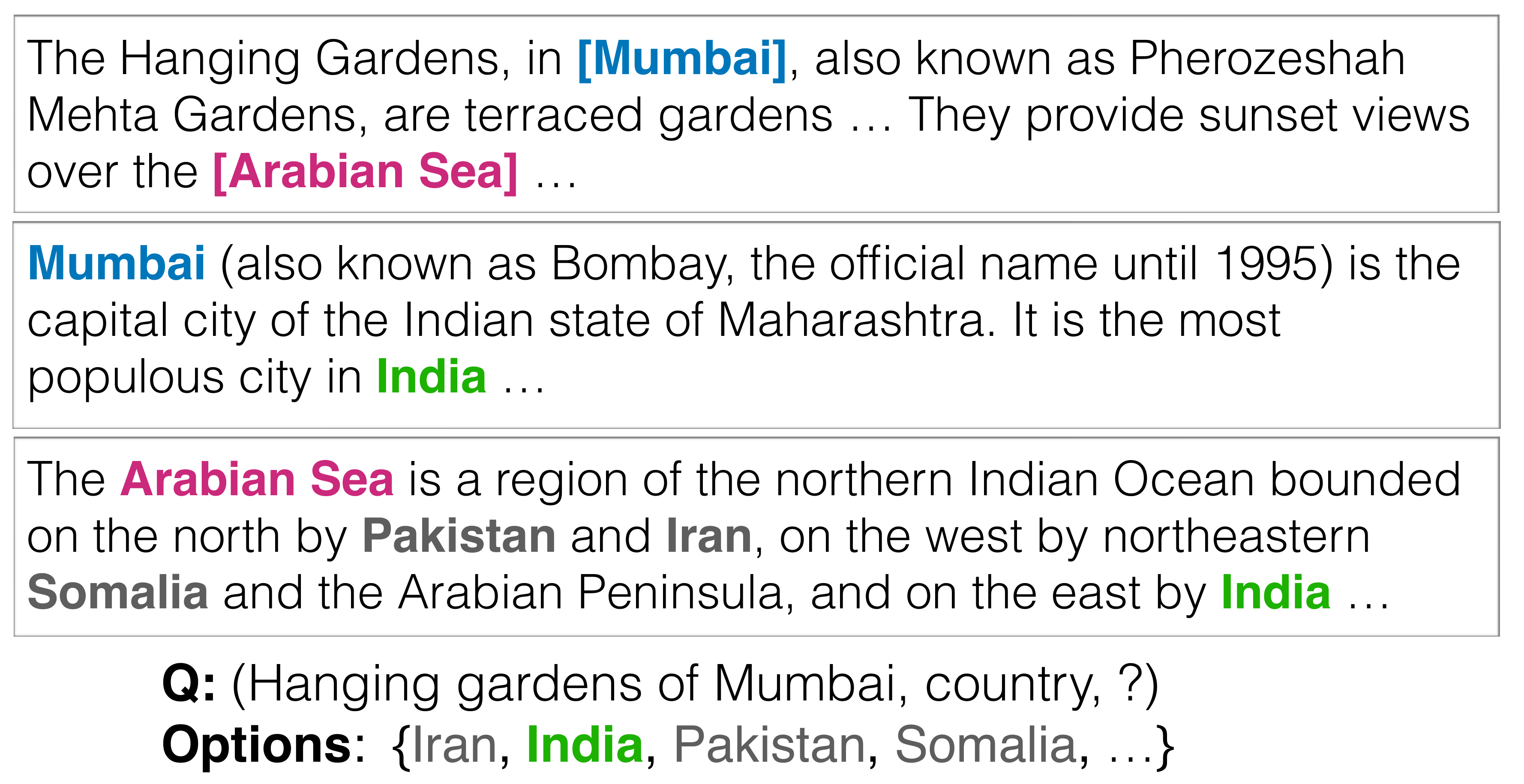}
    \caption{
        A sample from the \WikiHop dataset where it is necessary to combine information spread across multiple documents to infer the correct answer.
    }
    \label{fig:mumbai}
\end{figure}
Devising computer systems capable of answering questions about knowledge described using text has been a longstanding challenge in Natural Language Processing (NLP). 
Contemporary end-to-end Reading Comprehension (RC) methods can learn to extract the correct answer span within a given text and approach human-level performance~\cite{Kadlec2016ASR,Seo2016BidAF}.
However, for existing datasets, relevant information is often concentrated locally within a single sentence, emphasizing the role of locating, matching, and aligning information between query and support text.
For example, \newcite{Weissenborn2017fastQA} observed that a simple binary \emph{word-in-query} indicator feature boosted the relative accuracy of a baseline model by 27.9\%.

We argue that, in order to further the ability of machine comprehension methods to extract knowledge from text,
we must move beyond a scenario where relevant information is coherently and explicitly stated within a single document.
Methods with this capability would aid Information Extraction~(IE) applications, such as discovering drug-drug interactions ~\cite{gurulingappa12development} by connecting protein interactions reported across different publications.
They would also benefit search~\cite{Carpineto_2012_survey} and Question Answering~(QA) applications~\cite{Lin_2001_DIRT} where the required information cannot be found in a single location.

Figure~\ref{fig:mumbai} shows an example from \Wikipedia, where the goal is to identify the {\small \texttt{country}} property of the \emph{Hanging Gardens of Mumbai}.
This cannot be inferred solely from the article about them without additional background knowledge, as the answer is not stated explicitly.
However, several of the linked articles mention the correct answer \emph{India} (and other countries), but cover different topics~(e.g.~\emph{Mumbai}, \emph{Arabian Sea}, etc.).
Finding the answer requires \emph{multi-hop} reasoning:
figuring out that the \emph{Hanging Gardens} are located in \emph{Mumbai}, and then, from a second document, that \emph{Mumbai} is a city in \emph{India}.

We define a novel RC task in which a model should learn to answer queries by combining evidence stated across documents.
We introduce a methodology to induce datasets for this task and derive two datasets.
The first, \WikiHop, uses sets of \Wikipedia articles where answers to queries about specific properties of an entity cannot be located in the entity's article.
In the second dataset, \MedHop, the goal is to establish drug-drug interactions based on scientific findings about drugs and proteins and their interactions, found across multiple \MEDLINE abstracts.
For both datasets we draw upon existing Knowledge Bases (KBs), \Wikidata and \DrugBank, as ground truth, utilizing distant supervision~\cite{mintz2009distant} to induce the data -- similar to \newcite{hewlett2016_wikireading} and \newcite{Joshi_2017_TriviaQA}.

We establish that for 74.1\% and 68.0\% of the samples, the answer can be inferred from the given documents by a human annotator.
Still, constructing multi-document datasets is challenging; we encounter and prescribe remedies for several pitfalls associated with their assembly -- for example, spurious co-locations of answers and specific documents.

For both datasets we then establish several strong baselines and evaluate the performance of two previously proposed competitive RC models~\cite{Seo2016BidAF,Weissenborn2017fastQA}.
We find that one can integrate information across documents, but neither excels at selecting relevant information from a larger documents set, as their accuracy increases significantly when given only documents guaranteed to be relevant.
The best model reaches 54.5\% on an annotated test set, compared to human performance at 85.0\%, indicating ample room for improvement.

In summary, our key contributions are as follows:
Firstly, proposing a cross-document multi-step RC task, as well as a general dataset induction strategy.
Secondly, assembling two datasets from different domains and identifying dataset construction pitfalls and remedies.
Thirdly, establishing multiple baselines, including two recently proposed RC models, as well as analysing model behaviour in detail through ablation studies.

\section{Task and Dataset Construction Method}
\label{sec:datasets_abstract}
We will now formally define the multi-hop RC task, and a generic methodology to construct multi-hop RC datasets.
Later, in Sections~\ref{sec:wikihop} and~\ref{sec:medhop} we will demonstrate how this method is applied in practice by creating datasets for two different domains.
\paragraph{Task Formalization}
A model is given a query $q$, a set of supporting documents $S_q$, and a set of candidate answers $C_q$ -- all of which are mentioned in $S_q$.
The goal is to identify the correct answer $a^*~\in~C_q$ by drawing on the support documents $S_q$. 
Queries could potentially have several true answers when not constrained to rely on a specific set of support documents -- e.g.,\ queries about the parent of a certain individual.
However, in our setup each sample has only one true answer among $C_q$ and $S_q$.
Note that even though we will utilize background information during dataset assembly, such information will not be available to a model: 
the document set will be provided in random order and without any metadata. 
While certainly beneficial, this would distract from our goal of fostering end-to-end RC methods that infer facts by combining separate facts stated in text.

\paragraph{Dataset Assembly}
We assume that there exists a document corpus~$D$, together with a KB containing fact triples $(s, r, o)$ -- with subject entity~$s$, relation~$r$, and object entity~$o$.
For example, one such fact could be \texttt{\smaller (Hanging\_Gardens\_of\_Mumbai, country, India)}.
We start with individual KB facts and transform them into query-answer pairs by leaving the object slot empty, i.e.~$q=(s,r,?)$ and $a^{*}=o$.

\begin{figure}[t]
    \centering
    \includegraphics[width=0.6\columnwidth,trim={0 0.1cm 0 0.1cm},clip]{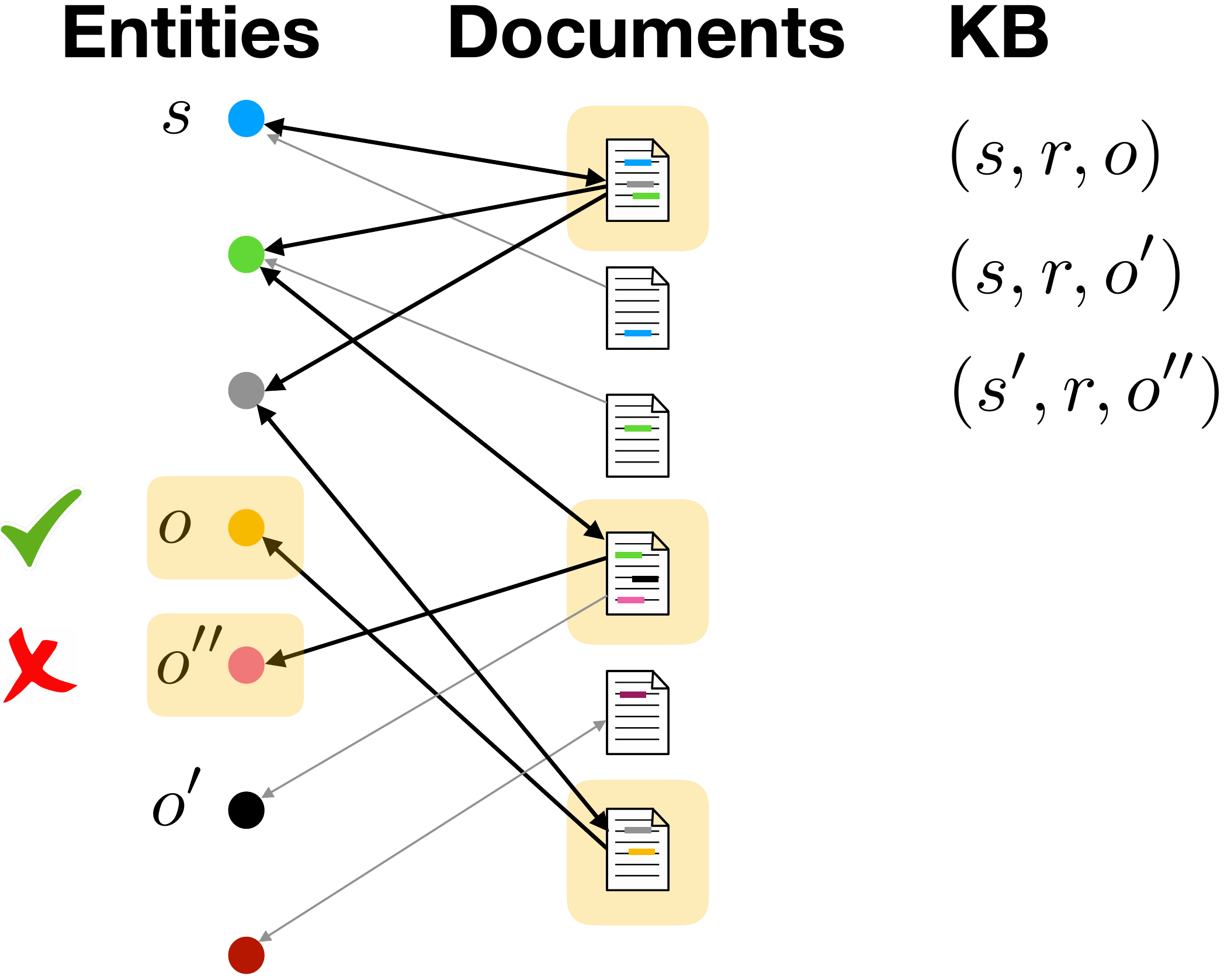}
    \caption{
        A bipartite graph connecting entities and documents mentioning them.
        Bold edges are those traversed for the first fact in the small KB on the right; 
        yellow highlighting indicates documents in $S_q$ and candidates in $C_q$.
        Check and cross indicate correct and false candidates.
    }
    \label{fig:bipartite}
\end{figure}

Next, we define a directed bipartite graph, where vertices on one side correspond to documents in $D$, and vertices on the other side are entities from the KB -- see Figure~\ref{fig:bipartite} for an example.
A document node~$d$ is connected to an entity $e$ if $e$ is mentioned in $d$, though there may be further constraints when defining the graph connectivity.
For a given $(q,a^*)$ pair, the candidates $C_q$ and support documents $S_q \subseteq D$ are identified by traversing the bipartite graph using breadth-first search; the documents visited will become the support documents $S_q$.

As the traversal starting point, we use the node belonging to the subject entity $s$ of the query $q$. 
As traversal end points, we use the set of all entity nodes that are type-consistent answers to $q$.%
\footnote{
    To determine entities which are type-consistent for a query~$q$, we consider all entities which are observed as object in a fact with $r$ as relation type -- including the correct answer.
}
Note that whenever there is another fact $(s,r,o')$ in the KB, i.e.\ a fact producing the same $q$ but with a different $a^*$, we will not include $o'$ into the set of end points for this sample.
This ensures that precisely one of the end points corresponds to a correct answer to $q$.\footnote{Here we rely on a closed-world assumption; that is, we assume that the facts in the KB state all true facts.}
%

%
When traversing the graph starting at $s$, several end points will be visited, though generally not all; those visited define the candidate set $C_q$.
%
If however the correct answer $a^*$ is not among them we discard the $(q,a^*)$ pair.
The documents visited to reach the end points will define the support document set $S_q$. 
%
That is, $S_q$ comprises chains of documents leading not only from the query subject to the correct answer, but also to type-consistent false candidates.

With this methodology, relevant textual evidence for $(q,a^*)$ will be spread across documents along the chain connecting $s$ and $a^*$ -- ensuring that multi-hop reasoning goes beyond resolving co-reference within a single document.
Note that including other type-consistent candidates alongside $a^*$ as end points in the graph traversal -- and thus into the support documents -- renders the task considerably more challenging~\cite{jia_2017_adversarial}.
Models could otherwise identify $a^*$ in the documents by simply relying on type-consistency heuristics. 
It is worth pointing out that by introducing alternative candidates we counterbalance a type-consistency bias, in contrast to \newcite{hermann2015teaching} and \newcite{Hill2015CBT} who instead rely on entity masking.
%
%

%

\section{\WikiHop}\label{sec:wikihop}
\Wikipedia contains an abundance of human-curated, multi-domain information and has several structured resources such as infoboxes and \Wikidata~\cite{vrandecic2012wikidata} associated with it.
\Wikipedia has thus been used for a wealth of research to build datasets posing queries about a single sentence~\cite{morales2016learning,levy2017zeroshot} or article~\cite{Yang2015_WikiQA,hewlett2016_wikireading,Rajpurkar2016_SQUAD}.
However, no attempt has been made to construct a cross-document multi-step RC dataset based on \Wikipedia.

A recently proposed RC dataset is \WikiReading~\cite{hewlett2016_wikireading}, where \Wikidata tuples {\smaller \texttt{(item, property, answer)}} are aligned with the \Wikipedia articles regarding their {\smaller \texttt{item}}.
The tuples define a slot filling task with the goal of predicting the {\smaller \texttt{answer}}, given an {\smaller \texttt{article}} and {\smaller \texttt{property}}.
One problem with using \WikiReading as an extractive RC dataset is that \WikiReadingUnseen of the samples do not state the answer explicitly in the given article~\cite{hewlett2016_wikireading}.
However, we observed that some of the articles accessible by following hyperlinks from the given article often state the answer, alongside other plausible candidates.

\subsection{Assembly}
We now apply the methodology from Section~\ref{sec:datasets_abstract} to create a multi-hop dataset with \Wikipedia as the document corpus and \Wikidata as structured knowledge triples.
In this setup, {\smaller \texttt{(item, property, answer)}} \Wikidata tuples correspond to $(s,r,o)$ triples, and the {\smaller \texttt{item}} and {\smaller \texttt{property}} of each sample together form our query $q$ -- e.g.,\ \emph{(Hanging Gardens of Mumbai, country, ?)}. 
Similar to \newcite{Yang2015_WikiQA} we only use the first paragraph of an article, as relevant information is more often stated in the beginning.
Starting with all samples in \WikiReading, we first remove samples where the {\smaller \texttt{answer}} is stated explicitly in the \Wikipedia article about the {\smaller \texttt{item}}.%
\footnote{
    We thus use a disjoint subset of \WikiReading compared to \newcite{levy2017zeroshot} to construct \WikiHop.
}

The bipartite graph is structured as follows: (1)~for edges from articles to entities: all articles mentioning an entity $e$ are connected to $e$; (2)~for edges from entities to articles: each entity $e$ is only connected to the \Wikipedia article about the entity.
Traversing the graph is then equivalent to iteratively following hyperlinks to new articles about the anchor text entities.

For a given query-answer pair, the {\smaller \texttt{item}} entity is chosen as the starting point for the graph traversal.
A traversal will always pass through the article about the {\small \texttt{item}}, since this is the only document connected from there. 
The end point set includes the correct {\texttt{\smaller answer}} alongside other type-consistent candidate expressions, which are determined by considering \emph{all} facts belonging to \WikiReading training samples, selecting those triples with the same {\smaller \texttt{property}} as in $q$ and keeping their {\smaller \texttt{answer}} expressions.
As an example, for the \Wikidata property {\smaller \texttt{country}}, this would be the set $\{\text{France}, \text{Russia}, ...\}$.
We executed graph traversal up to a maximum chain length of 3 documents.
To not pose unreasonable computational constraints, samples with more than 64 different support documents or 100 candidates are removed, discarding $\approx$1\% of the samples.

\subsection{Mitigating Dataset Biases}
\label{sec:mitigating}
Dataset creation is always fraught with the risk of inducing unintended errors and biases~\cite{chen_2016_thorough,schwartz2017effect}.
As \newcite{hewlett2016_wikireading} only carried out limited analysis of their \WikiReading dataset, we present an analysis of the downstream effects we observe on \WikiHop.

\paragraph{Candidate Frequency Imbalance}
A first observation is that there is a significant bias in the answer distribution of \WikiReading.
For example, in the majority of the samples the property {\small \texttt{country}} has the \emph{United States of America} as the answer.
A simple majority class baseline would thus prove successful, but would tell us little about multi-hop reasoning.
To combat this issue, we subsampled the dataset to ensure that samples of any one particular answer candidate make up no more than $0.1\%$ of the dataset, and omitted articles about the \emph{United States}.

\paragraph{Document-Answer Correlations}
\label{sec:document_answer}
A problem unique to our multi-document setting is the possibility of spurious correlations between candidates and documents induced by the graph traversal method. 
In fact, if we were \emph{not} to address this issue, a model designed to exploit these regularities could achieve 74.6\% accuracy (detailed in Section~\ref{sec:experiments}). 

Concretely, we observed that certain documents frequently co-occur with the correct answer, independently of the query.
For example, if the article about \emph{London} is present in $S_q$, the answer is likely to be the \emph{United Kingdom}, independent of the query type or entity in question.
Appendix~\ref{appendix:cue-examples} contains a list with several additional examples.

We designed a statistic to measure this effect and then used it to sub-sample the dataset. 
The statistic counts how often a candidate $c$ is observed as the correct answer when a certain document is present in $S_q$ across training set samples.
More formally, for a given document $d$ and answer candidate $c$, let $\text{\emph{cooccurrence}}(d,c)$ denote the total count of how often $d$ co-occurs with $c$ in a sample where $c$ is also the correct answer.
We use this statistic to filter the dataset, by discarding samples with at least one document-candidate pair $(d,c)$ for which $\text{\emph{cooccurrence}}(d,c)>20$. 
\section{\MedHop}\label{sec:medhop}
Following the same general methodology, we next construct a second dataset for the domain of molecular biology -- a field that has been undergoing exponential growth in the number of publications~\cite{cohen2004natural}.
The promise of applying NLP methods to cope with this increase has led to research efforts in IE~\cite{hirschman2005overview,kim2011overview} and QA for biomedical text~\cite{hersh2007trec,nentidis2017results}.
There are a plethora of manually curated structured resources~\cite{ashburner2000gene,uniprot2016uniprot} which can either serve as ground truth or to induce training data using distant supervision~\cite{craven1999constructing,bobic2012improving}.
Existing RC datasets are either severely limited in size~\cite{hersh2007trec} or cover a very diverse set of query types~\cite{nentidis2017results}, complicating the application of neural models that have seen successes for other domains~\cite{wiese2017wiese}.

A task that has received significant attention is detecting Drug-Drug Interactions~(DDIs).
Existing DDI efforts have focused on explicit mentions of interactions in single sentences~\cite{gurulingappa12development,percha2012discovery,segurabedmar2013ddi}.
However, as shown by \newcite{Peng2017_cross_sentence}, cross-sentence relation extraction increases the number of available relations.
It is thus likely that cross-document interactions would further improve recall,  which is of particular importance considering interactions that are never stated explicitly -- but rather need to be inferred from separate pieces of evidence.
The promise of multi-hop methods is finding and combining individual observations that can suggest previously unobserved DDIs, aiding the process of making scientific discoveries, yet not directly from experiments, but by inferring them from established public knowledge~\cite{swanson1986undiscovered}.

\begin{figure}[ht]
	\centering
	\includegraphics[width=1.0\columnwidth,trim={0.1cm 13.2cm 0.1cm 0.1cm},clip]{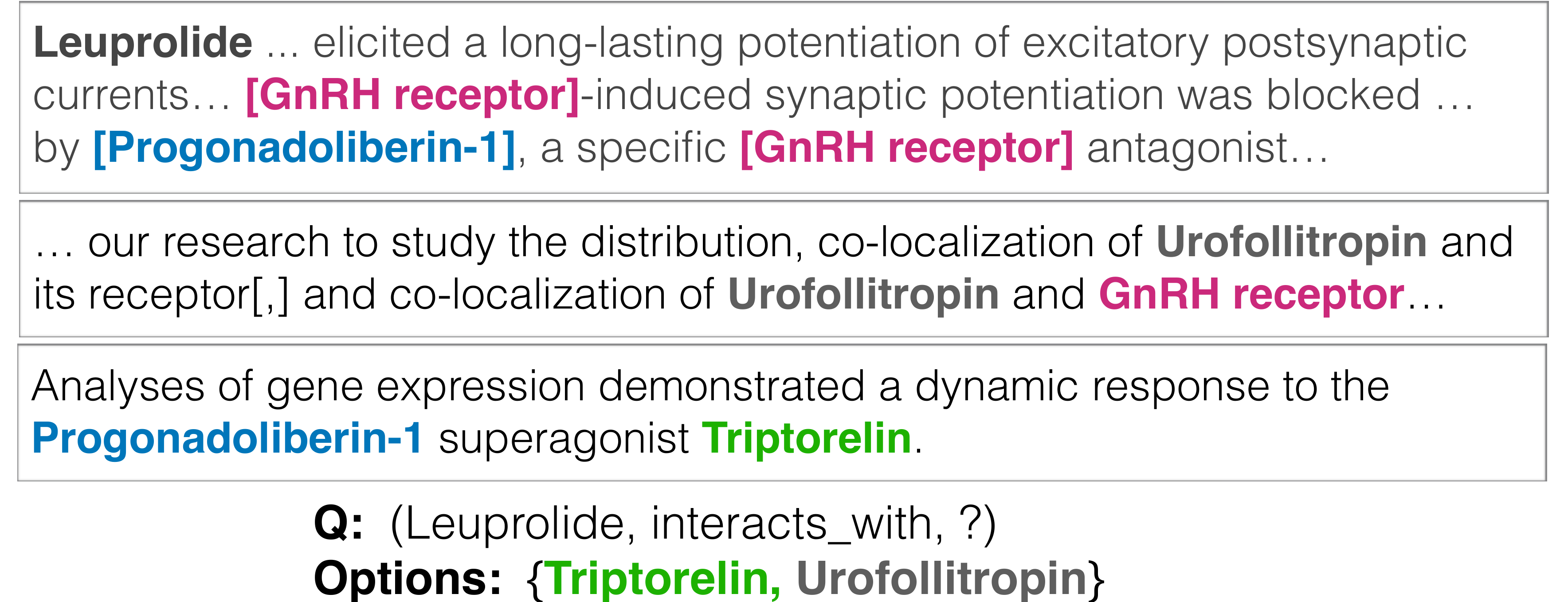}
	\caption{
	    A sample from the \MedHop dataset.
	}
    \label{fig:medhop}
\end{figure}

DDIs are caused by Protein-Protein Interaction (PPI) chains, forming biomedical pathways.
If we consider PPI chains across documents, we find examples like in Figure~\ref{fig:medhop}.
Here the first document states that the drug {\smaller \texttt{Leuprolide}} causes {\smaller \texttt{GnRH receptor}}-induced synaptic potentiations, which can be blocked by the protein {\smaller \texttt{Progonadoliberin-1}}.
The last document states that another drug, {\smaller \texttt{Triptorelin}}, is a superagonist of the same protein.
It is therefore likely to affect the potency of {\smaller \texttt{Leuprolide}}, describing a way in which the two drugs interact.
Besides the true interaction there is also a false candidate {\smaller \texttt{Urofollitropin}} for which, although mentioned together with {\smaller \texttt{GnRH receptor}} within one document, there is no textual evidence indicating interactions with {\smaller \texttt{Leuprolide}}.

\subsection{Assembly}
We construct \MedHop using \DrugBank~\cite{law2014drugbank} as structured knowledge resource and research paper abstracts from \MEDLINE as documents. 
There is only one relation type for \DrugBank facts, {\smaller \texttt{interacts\_with}}, that connects pairs of drugs -- an example of a \MedHop query would thus be \texttt{\smaller (Leuprolide, interacts\_with, ?)}.
We start by processing the 2016 \MEDLINE release using the preprocessing pipeline employed for the BioNLP 2011 Shared Task~\cite{stenetorp2011supporting}.
We restrict the set of entities in the bipartite graph to drugs in \DrugBank and human proteins in \SwissProt~\cite{bairoch2004swissprot}.
That is, the graph has drugs and proteins on one side, and \MEDLINE abstracts on the other.

The edge structure is as follows:
(1)~There is an edge from a document to all proteins mentioned in it.
(2)~There is an edge between a document and a drug, if this document also mentions a protein known to be a target for the drug according to \DrugBank.
This edge is bidirectional, i.e.\ it can be traversed both ways, since there is no canonical document describing each drug -- thus one can ``hop'' to any document mentioning the drug and its target.
(3)~There is an edge from a protein $p$ to a document mentioning $p$, but only if the document also mentions another protein $p'$ which is known to interact with $p$ according to \textsc{Reactome}~\cite{fabregat2016reactome}.
Given our distant supervision assumption, these additionally constraining requirements err on the side of precision.

As a mention, similar to \newcite{percha2012discovery}, we consider any exact match of a name variant of a drug or human protein in \DrugBank or \SwissProt.
For a given DDI \texttt{\smaller(drug$_1$,  interacts\_with, drug$_2$)}, we then select \texttt{\smaller drug$_1$} as the starting point for the graph traversal.
As possible end points, we consider any other drug, apart from \texttt{\smaller drug$_1$} and those interacting with \texttt{\smaller drug$_1$} other than \texttt{\smaller drug$_2$}.
Similar to \WikiHop, we exclude samples with more than 64 support documents and impose a maximum document length of 300 tokens plus title.

\paragraph{Document Sub-sampling}
The bipartite graph for \MedHop is orders of magnitude more densely connected than for \WikiHop.
This can lead to potentially large support document sets $S_q$, to a degree where it becomes computationally infeasible for a majority of existing RC models.
After the traversal has finished, we subsample documents by first adding a set of documents that connects the drug in the query with its answer.
We then iteratively add documents to connect alternative candidates until we reach the limit of 64 documents -- while ensuring that all candidates have the same number of paths through the bipartite graph.

\paragraph{Mitigating Candidate Frequency Imbalance}
Some drugs interact with more drugs than others -- \textit{Aspirin} for example interacts with 743 other drugs, but \textit{Isotretinoin} with only 34.
This leads to similar candidate frequency imbalance issues as with {\WikiHop} -- but due to its smaller size \MedHop is difficult to sub-sample.
Nevertheless we can successfully combat this issue by masking entity names, detailed in Section~\ref{sec:masking}.

\section{Dataset Analysis}
Table~\ref{tbl:dataset_sizes} shows the dataset sizes.
Note that \WikiHop inherits the train, development, and test set splits from {\WikiReading} -- i.e., the full dataset creation, filtering, and sub-sampling pipeline is executed on each set individually.
Also note that sub-sampling according to document-answer correlation significantly reduces the size of \WikiHop from ${\approx}528$K training samples to ${\approx}44$K.
While in terms of samples, both \WikiHop and \MedHop are smaller than other large-scale RC datasets, such as \emph{SQuAD} and \WikiReading, the supervised learning signal available per sample is arguably greater.
One could, for example, re-frame the task as binary path classification: given two entities and a document path connecting them, determine whether a given relation holds.
For such a case, \WikiHop and \MedHop would have more than 1M and 150K paths to be classified, respectively. Instead, in our formulation, this corresponds to each single sample containing the supervised learning signal from an average of 19.5 and 59.8 unique document paths.

\begin{table}
    \centering
    \begin{tabular}{lrrrr}
                    & Train     & Dev   & Test  & Total     \\
        \toprule
        \WikiHop    & 43,738    & 5,129 & 2,451 & 51,318    \\
        \MedHop     &  1,620    &   342 &   546 &  2,508     \\
    \end{tabular}
    \caption{
        Dataset sizes for our respective datasets.
    }
    \label{tbl:dataset_sizes}
\end{table}
\begin{table}
    \centering
    \begin{tabular}{lrrrr}
                            & min       & max   & avg   &  median       \\
        \toprule
        \# cand.  -- WH     & 2         & 79    & 19.8  & 14    \\
        \# docs.  -- WH     & 3         & 63    & 13.7  & 11    \\
        \# tok/doc -- WH & 4         & 2,046  & 100.4 & 91    \\
        \midrule
        \# cand.  -- MH     & 2         & 9     & 8.9   & 9      \\
        \# docs.  -- MH     & 5         & 64    & 36.4  & 29    \\
        \# tok/doc  -- MH   & 5         & 458    & 253.9  & 264    \\
    \end{tabular}
    \caption{
        Candidates and documents per sample and document length statistics. WH: \WikiHop; MH: \MedHop.
    }
    \label{tbl:dataset_minmaxmean}
\end{table}

Table~\ref{tbl:dataset_minmaxmean} shows statistics on the number of candidates and documents per sample on the respective training sets.
For \MedHop, the majority of samples have 9 candidates, due to the way documents are selected up until a maximum of 64 documents is reached.
Few samples have less than 9 candidates, and samples would have far more false candidates if more than 64 support documents were included.
The number of query types in \WikiHop is 277, whereas in \MedHop there is only one: {\smaller \texttt{interacts\_with}}.

\subsection{Qualitative Analysis}
\label{sec:qualitative}
To establish the quality of the data and analyze potential distant supervision errors, we sampled and annotated 100 samples from each development set.

\paragraph{\WikiHop}
Table~\ref{tbl:wikihop_qualitative} lists characteristics along with the proportion of samples that exhibit them.
For 45\%, the true answer either uniquely follows from multiple texts directly or is suggested as likely.
For 26\%, more than one candidate is plausibly supported by the documents, including the correct answer.
This is often due to hypernymy, where the appropriate level of granularity for the answer is difficult to predict --  e.g.\
{\smaller \texttt{(west~suffolk, administrative\_entity, ?)}} with candidates {\smaller \texttt{suffolk}} and {\smaller \texttt{england}}.
This is a direct consequence of including type-consistent false answer candidates from \Wikidata, which can lead to questions with several true answers.
For 9\% of the cases a single document suffices; these samples contain a document that states enough information about {\smaller \texttt{item}} and {\smaller \texttt{answer}} together. 
For example, the query {\smaller \texttt{(Louis Auguste, father, ?)}} has the correct answer {\smaller \texttt{Louis XIV of France}}, and {\smaller \texttt{French king Louis XIV}} is mentioned within the same document as {\smaller \texttt{Louis Auguste}}.
Finally, although our task is significantly more complex than most previous tasks where distant supervision has been applied, the distant supervision assumption is only violated for 20\% of the samples -- a proportion similar to previous work~\cite{Riedel_2010_modeling}.
These cases can either be due to conflicting information between \Wikidata and \Wikipedia~(8\%), e.g.\ when the date of birth for a person differs between \Wikidata and what is stated in the \Wikipedia article, or because the answer is consistent but cannot be inferred from the support documents~(12\%).
When answering 100 questions, the annotator knew the answer prior to reading the documents for 9\%, and produced the correct answer after reading the document sets for 74\% of the cases.
On 100 questions of a validated portion of the Dev set (see Section~\ref{sec:test_verification}), 85\% accuracy was reached.

\paragraph{\MedHop}
Since both document complexity and number of documents per sample were significantly larger compared to \WikiHop, (see Figure~\ref{fig:num_supports} in Appendix~\ref{appendix:plots}) it was not feasible to ask an annotator to read \emph{all} support documents for 100 samples.
%
We opted to verify the dataset quality by providing only the subset of documents relevant to support the correct answer, i.e., those traversed along the path reaching the answer.
The annotator was asked if the answer to the query \textit{``follows''}, \textit{``is likely''}, or \textit{``does not follow''}, given the relevant documents.
68\% of the cases were considered as \textit{``follows''} or as \textit{``is likely''}.
%
%
The majority of cases violating the distant supervision assumption were due to lacking a necessary PPI in one of the connecting documents.

\begin{table}[t]
    \centering
    \begin{tabular}{lr}
        Unique multi-step answer.  & 36\%                          \\
        Likely multi-step unique answer.                         &  9\%                          \\
        Multiple plausible answers.          & 15\%                          \\
        Ambiguity due to hypernymy.                              & 11\%                          \\
        Only single document required.           &  9\%                          \\
        \midrule
        Answer does not follow.                              & 12\%                          \\
        \Wikidata/\Wikipedia discrepancy.     &  8\%                          \\
    \end{tabular}
    \caption{
       Qualitiative analysis of \WikiHop samples.
    }
    \label{tbl:wikihop_qualitative}
\end{table}

\subsection{Crowdsourced Human Annotation}\label{sec:crowdsourcing}
We asked human annotators on \emph{Amazon Mechanical Turk} to evaluate samples of the \WikiHop development set.
Similar to our qualitative analysis of \MedHop, annotators were shown the query-answer pair as a fact and the chain of relevant documents leading to the answer.
They were then instructed to answer (1)~whether they knew the fact before; (2)~whether the fact follows from the texts (with options \emph{``fact follows''}, \emph{``fact is likely''}, and \emph{``fact does not follow''}); and~(3); whether a single or several of the documents are required.
Each sample was shown to three annotators and a majority vote was used to aggregate the annotations.
Annotators were familiar with the fact 4.6\% of the time; prior knowledge of the fact is thus not likely to be a confounding effect on the other judgments.
Inter-annotator agreement as measured by Fleiss' kappa is 0.253 in~(2), and 0.281 in~(3) -- indicating a fair overall agreement, according to \newcite{landis1977measurement}.
Overall, 9.5\% of samples have no clear majority in~(2).

Among samples with a majority judgment, 59.8\% are cases where the fact \emph{``follows''}, for 14.2\% the fact is judged as \emph{``likely''}, and as \emph{``not follow''} for 25.9\%.
This again provides good justification for the distant supervision strategy.

Among the samples with a majority vote for~(2) of either \emph{``follows''} or \emph{``likely''}, 55.9\% were marked with a majority vote as requiring multiple documents to infer the fact, and 44.1\% as requiring only a single document.
The latter number is larger than initially expected, given the construction of samples through graph traversal.
However, when inspecting cases judged as \emph{``single''} more closely, we observed that many indeed provide a clear hint about the correct answer within one document, but without stating it explicitly.
For example, for the fact \texttt{\smaller(witold cichy, country\_of\_citizenship, poland)} with documents \emph{$d_1$: Witold Cichy (born March 15, 1986 in Wodzisław Śląski) is a Polish footballer[...]} and \emph{$d_2$: Wodzisław Śląski[...] is a town in Silesian Voivodeship, southern Poland[...]}, the information provided in $d_1$ suffices for a human given the background knowledge that \emph{Polish} is an attribute related to \emph{Poland}, removing the need for $d_2$ to infer the answer.

\subsection{Validated Test Sets}\label{sec:test_verification}
While training models on distantly supervised data is useful, one should ideally evaluate methods on a manually validated test set.
We thus identified subsets of the respective test sets for which the correct answer can be inferred from the text.
This is in contrast to prior work such as \newcite{hermann2015teaching}, \newcite{Hill2015CBT}, and \newcite{hewlett2016_wikireading}, who evaluate only on distantly supervised samples.
For \WikiHop, we applied the same annotation strategy as described in Section~\ref{sec:crowdsourcing}.
The validated test set consists of those samples labeled by a majority of annotators (at least 2 of 3) as \emph{``follows''}, and requiring  \emph{``multiple''} documents.
While desirable, crowdsourcing is not feasible for \MedHop since it requires specialist knowledge.
In addition, the number of document paths is $\approx$3x larger, which along with the complexity of the documents greatly increases the annotation time.
We thus manually annotated 20\% of the \MedHop test set and identified the samples for which the text implies the correct answer and where multiple documents are required.

\section{Experiments}\label{sec:experiments}
This section describes experiments on \WikiHop and \MedHop with the goal of establishing the performance of several baseline models, including recent neural RC models.
We empirically demonstrate the importance of mitigating dataset biases, probe whether multi-step behavior is beneficial for solving the task, and investigate if RC models can learn to perform lexical abstraction.
Training will be conducted on the respective training sets, and evaluation on both the full test set and validated portion~(Section~\ref{sec:test_verification}) allowing for a comparison between the two.

\subsection{Models}
\paragraph{Random}
Selects a random candidate; note that the number of candidates differs between samples.

\paragraph{Max-mention}
Predicts the most frequently mentioned candidate in the support documents $S_q$ of a sample -- randomly breaking ties. 

\paragraph{Majority-candidate-per-query-type}Predicts the candidate $c\in C_q$ that was most frequently observed as the true answer in the training set, given the query type of $q$.
For \WikiHop, the query type is the property $p$ of the query; for \MedHop there is only the single query type -- {\smaller \texttt{interacts\_with}}.

\paragraph{TF-IDF}
Retrieval-based models are known to be strong QA baselines if candidate answers are provided~\cite{clar_2016_combining,welbl2017crowdsourcing}.
They search for individual documents based on keywords in the question, but typically do not combine information across documents.
The purpose of this baseline is to see if it is possible to identify the correct answer from a single document alone through lexical correlations.
The model forms its prediction as follows:
For each candidate $c$, the concatenation of the query $q$ with $c$ is fed as an \emph{OR} query into the \emph{whoosh} text retrieval engine.\footnote{
    \url{https://pypi.python.org/pypi/Whoosh/}
}
It then predicts the candidate with the highest TF-IDF similarity score:
\begin{equation}
    ~\argmax_{c \in C_{q}} \large[ \max_{s \in S_{q}} (\emph{TF-IDF}(q+c,s)) \large]~~
\end{equation}

\paragraph{Document-cue} 
During dataset construction we observed that certain document-answer pairs appear more frequently than others, to the effect that the correct candidate is often indicated solely by the presence of certain documents in $S_q$.
This baseline captures how easy it is for a model to exploit these informative document-answer co-occurrences.
It predicts the candidate with highest score across~$C_q$:
\begin{equation}
    \label{eq:nda}
    \argmax_{c \in C_q}\large[\max_{d \in S_q} ( \text{\emph{cooccurrence}}(d,c) ) \large]~
\end{equation}

\paragraph{Extractive RC models: FastQA and BiDAF}
In our experiments we evaluate two recently proposed \emph{LSTM}-based extractive QA models: the Bidirectional Attention Flow model (\emph{BiDAF}, \newcite{Seo2016BidAF}), and \emph{FastQA}~\cite{Weissenborn2017fastQA}, which have shown a robust performance across several datasets.
These models predict an answer span within a \emph{single} document.
We adapt them to a multi-document setting by sequentially concatenating all $d\in S_q$ in random order into a superdocument, adding document separator tokens.
During training, the first answer mention in the concatenated document serves as the gold span.\footnote{
    We also tested assigning the gold span randomly to any one of the mention of the answer, with insignificant changes.
}
At test time, we measured accuracy based on the exact match between the prediction and answer, both lowercased, after removing articles, trailing white spaces and punctuation, in the same way as \newcite{Rajpurkar2016_SQUAD}.
To rule out any signal stemming from the order of documents in the superdocument, this order is randomized both at training and test time.
In a preliminary experiment we also trained models using different random document order permutations, but found that performance did not change significantly.

For \emph{BiDAF}, the default hyperparameters from the implementation of \newcite{Seo2016BidAF} are used, with pretrained GloVe~\cite{pennington_2014_glove} embeddings.
However, we restrict the maximum document length to 8,192 tokens and hidden size to 20, and train for 5,000 iterations with batchsize 16 in order to fit the model into memory.
\footnote{
    The superdocument has a larger number of tokens compared to e.g.~\emph{SQuAD}, thus the additional memory requirements.
}
For \emph{FastQA} we use the implementation provided by the authors, also with pre-trained GloVe embeddings, no character-embeddings, no maximum support length, hidden size 50, and batch size 64 for 50 epochs.

While \emph{BiDAF} and \emph{FastQA} were initially developed and tested on single-hop RC datasets, their usage of bidirectional LSTMs and attention over the full sequence theoretically gives them the capacity to integrate information from different locations in the (super-)document.
In addition, \emph{BiDAF} employs iterative conditioning across multiple layers, potentially making it even better suited to integrate information found across the sequence.

\subsection{Lexical Abstraction: Candidate Masking}
\label{sec:masking}
The presence of lexical regularities among answers is a problem in RC dataset assembly -- a phenomenon already observed by \newcite{hermann2015teaching}.
When comprehending a text, the correct answer should become clear from its context -- rather than from an intrinsic property of the answer expression.
To evaluate the ability of models to rely on context alone, we created \emph{masked} versions of the datasets: we replace any candidate expression randomly using 100 unique placeholder tokens, e.g.\ \emph{``Mumbai is the most populous city in \textsc{MASK7}."}
Masking is consistent within one sample, but generally different for the same expression across samples. 
This not only removes answer frequency cues, it also removes statistical correlations between frequent answer strings and support documents.
Models consequently cannot base their prediction on intrinsic properties of the answer expression, but have to rely on the context surrounding the mentions.

\begin{table}[t]
    \begin{center}
        \begin{tabular}{lrr}
            Model               & Unfiltered    &  Filtered         \\
            \toprule
            Document-cue        & 74.6        & 36.7           \\
            Maj. candidate      & 41.2        & 38.8           \\
            TF-IDF              & 43.8        & 25.6           \\
            \midrule
            Train set size      & 527,773     & 43,738              
        \end{tabular}
    \end{center}
    \caption{
        Accuracy comparison for simple baseline models on \WikiHop \emph{before} and \emph{after} filtering.
    }
    \label{tab:wh3_subsampling_effect}
\end{table}

\begin{table*}[ht]
    \begin{center}
        \begin{tabular} {@{\extracolsep{2pt}}lrrrrrrrr@{}}
                 & \multicolumn{4}{c}{\textbf{\WikiHop}} & \multicolumn{4}{c}{\textbf{\MedHop}} \\ 
                 & \multicolumn{2}{c}{\textbf{standard}} &\multicolumn{2}{c}{\textbf{masked}} & \multicolumn{2}{c}{\textbf{standard}} & \multicolumn{2}{c}{\textbf{masked}}\\
                 \cline{2-3} \cline{4-5} \cline{6-7} \cline{8-9}
            Model                    & test       &   test* & test & test*    & test            & test*         & test     & test* \\
            \toprule
            Random                   & 11.5     & 12.2  & 12.2  & 13.0          & 13.9          & 20.4          & 14.1        & 22.4    \\
            Max-mention              & 10.6     & 15.9  & 13.9  & 20.1          &  9.5          & 16.3          &  9.2        & 16.3     \\
         Majority-candidate-per-query-type     & 38.8     & 44.2  & 12.0  & 13.7          & \textbf{58.4} & \textbf{67.3} & 10.4        & 6.1     \\
            TF-IDF                   & 25.6     & 36.7  & 14.4  & 24.2          &  9.0          & 14.3          &  8.8        & 14.3     \\
            Document-cue             & 36.7     & 41.7  &  7.4  & 20.3          & 44.9          & 53.1          & 15.2        & 16.3     \\
            \midrule
            FastQA                   & 25.7     & 27.2  & 35.8  & 38.0          & 23.1          & 24.5          & 31.3        & 30.6     \\
            BiDAF       &\textbf{42.9}  &\textbf{49.7}  &\textbf{54.5}& \textbf{59.8}   & 47.8  & 61.2          &\textbf{33.7}&\textbf{42.9}     \\

        \end{tabular}
    \end{center}
    \caption{
        Test accuracies for the \WikiHop and \MedHop datasets, both in standard (unmasked) and masked setup. Columns marked with asterisk are for the validated portion of the dataset.
    }
    \label{tab:main_results}
\end{table*}

\begin{table*}[ht]
    \begin{center}
        \begin{tabular}{@{\extracolsep{2pt}} l r r r r r r r r @{}}
          & \multicolumn{4}{c}{\textbf{\WikiHop}} & \multicolumn{4}{c}{\textbf{\MedHop}} \\
        & \multicolumn{2}{c}{\textbf{standard}} & \multicolumn{2}{c}{\textbf{gold chain}} &\multicolumn{2}{c}{\textbf{standard}} & \multicolumn{2}{c}{\textbf{gold chain}}\\
                 \cline{2-3} \cline{4-5} \cline{6-7} \cline{8-9}
         Model          & test          & test*     & test          & test*     &test           & test*         & test          & test*     \\
         \toprule
         \BiDAF         &  42.9         & 49.7      & 57.9          &  63.4     & \textbf{47.8} & \textbf{61.2} & 86.4          &  89.8    \\
         \BiDAF mask  &\textbf{54.5}& \textbf{59.8} &\textbf{81.2}  &\textbf{85.7}  &  33.7     & 42.9          & \textbf{99.3} & \textbf{100.0}\\
         \midrule
         \FastQA        &  25.7         & 27.2      & 44.5          & 53.5          &23.1       & 24.5          & 54.6          & 59.2      \\
         \FastQA mask   &  35.8         & 38.0      & 65.3          & 70.0          &31.3       & 30.6          & 51.8          & 55.1      \\
        \end{tabular}
    \end{center}
    \caption{
        Test accuracy comparison when only using documents leading to the correct answer (gold chain). Columns with asterisk hold results for the validated samples.
    }
    \label{tbl:oracle}
\end{table*}

\subsection{Results and Discussion}

Table~\ref{tab:main_results} shows the experimental outcomes for \WikiHop and \MedHop, together with results for the \emph{masked} setting; we will first discuss the former.
A first observation is that candidate mention frequency does not produce better predictions than a random guess.
Predicting the answer most frequently observed at training time achieves strong results: as much as 38.8\% / 44.2\% and 58.4\% / 67.3\% on the two datasets, for the full and validated test sets respectively.
That is, a simple frequency statistic together with answer type constraints alone is a relatively strong predictor, and the strongest overall for the \textit{``unmasked''} version of \MedHop.

The TF-IDF retrieval baseline clearly performs better than random for \WikiHop, but is not very strong overall.
That is, the question tokens are helpful to detect relevant documents, but exploiting only this information compares poorly to the other baselines.
On the other hand, as no co-mention of an interacting drug pair occurs within any single document in \MedHop, the TF-IDF baseline performs worse than random.
We conclude that lexical matching with a single support document is not enough to build a strong predictive model for both datasets.

The \emph{Document-cue} baseline can predict more than a third of the samples correctly, for both datasets, even after sub-sampling frequent document-answer pairs for \WikiHop.
The relative strength of this and other baselines proves to be an important issue when designing multi-hop datasets, which we addressed through the measures described in Section~\ref{sec:mitigating}.
In Table~\ref{tab:wh3_subsampling_effect} we compare the two relevant baselines on \WikiHop before and after applying filtering measures.
The absolute strength of these baselines before filtering shows how vital addressing this issue is: 74.6\% accuracy could be reached through exploiting the $\text{\emph{cooccurrence}}(d,c)$ statistic alone.
This underlines the paramount importance of investigating and addressing dataset biases that otherwise would confound seemingly strong RC model performance.
The relative drop demonstrates that the measures undertaken successfully mitigate the issue.
A downside to aggressive filtering is a significantly reduced dataset size, rendering it infeasible for smaller datasets like \MedHop.

Among the two neural models, \emph{BiDAF} is overall strongest across both datasets -- this is in contrast to the reported results for SQuAD where their performance is nearly indistinguishable.
This is possibly due to the iterative latent interactions in the \emph{BiDAF} architecture: we hypothesize that these are of increased importance for our task, where information is distributed across documents.
It is worth emphasizing that unlike the other baselines, both \emph{FastQA} and \emph{BiDAF} predict the answer by extracting a span from the support documents without relying on the candidate options $C_q$.

In the \emph{masked} setup all baseline models reliant on lexical cues fail in the face of the randomized answer expressions, since the same answer option has different placeholders in different samples.
Especially on \MedHop, where dataset sub-sampling is not a viable option, masking proves to be a valuable alternative, effectively circumventing spurious statistical correlations that RC models can learn to exploit.

Both neural RC models are able to largely retain or even improve their strong performance when answers are masked: they are able to leverage the textual context of the candidate expressions.
To understand differences in model behavior between \WikiHop and \MedHop, it is worth noting that drug mentions in \MedHop are normalized to a unique single-word identifier, and performance drops under masking.
In contrast, for the open-domain setting of \WikiHop, a reduction of the answer vocabulary to 100 random single-token \emph{mask} expressions clearly helps the model in selecting a candidate span, compared to the multi-token candidate expressions in the unmasked setting.
Overall, although both neural RC models clearly outperform the other baselines, they still have large room for improvement compared to human performance at 74\% / 85\% for \WikiHop.

Comparing results on the full and validated test sets, we observe that the results consistently improve on the validated sets.
This suggests that the training set contains the signal necessary to make inference on valid samples at test time, and that noisy samples are harder to predict.

\begin{table}
    \begin{center}
        \begin{tabular}{@{\extracolsep{4pt}} l l l l l @{}}
          & \multicolumn{2}{c}{\textbf{\WikiHop}} & \multicolumn{2}{c}{\textbf{\MedHop}} \\
                          \cline{2-3} \cline{4-5}
                                &   test      & test*       & test      & test*  \\
         \toprule
         \BiDAF                 &  54.5   & 59.8            & 33.7    & 42.9    \\
         \BiDAF rem             &  44.6   & 57.7            & 30.4    & 36.7    \\
         \midrule
         \FastQA                &  35.8   & 38.0            & 31.3    & 30.6    \\
         \FastQA rem            &  38.0   & 41.2            & 28.6    & 24.5    \\
        \end{tabular}
    \end{center}
    \caption{
        Test accuracy (masked) when only documents containing answer candidates are given (\emph{rem}).
    }
    \label{tbl:doc-rem}
\end{table}
\subsection{Using only relevant documents}
We conducted further experiments to examine the RC models when presented with only the relevant documents in $S_q$, i.e.,\ the chain of documents leading to the correct answer.
This allows us to investigate the hypothetical performance of the models if they were able to select and read only relevant documents: Table~\ref{tbl:oracle} summarizes these results.
Models improve greatly in this \emph{gold chain} setup, with up to 81.2\% / 85.7\% on \WikiHop in the masked setting for \emph{BiDAF}.
This demonstrates that RC models are capable of identifying the answer when few or no plausible false candidates are mentioned, which is particularly evident for \MedHop, where documents tend to discuss only single drug candidates.
In the \emph{masked} gold chain setup, models can then pick up on what the masking template looks like and achieve almost perfect scores.
Conversely, these results also show that the models' answer selection process is not robust to the introduction of unrelated documents with type-consistent candidates.
This indicates that learning to intelligently select relevant documents before RC may be among the most promising directions for future model development.

\subsection{Removing relevant documents}
To investigate if the neural RC models can draw upon information requiring multi-step inference we designed an experiment where we discard all documents that do not contain candidate mentions, including the first documents traversed.
Table~\ref{tbl:doc-rem} shows the results: we can observe that performance drops across the board for \BiDAF.
There is a significant drop of 3.3\%/6.2\% on \MedHop, and 10.0\%/2.1\% on \WikiHop, demonstrating that \BiDAF, is able to leverage cross-document information.
\FastQA shows a slight increase of 2.2\%/3.2\% for \WikiHop and a decrease of 2.7\%/4.1\% on \MedHop.
While inconclusive, it is clear that \FastQA with fewer latent interactions than \BiDAF has problems integrating cross-document information.

\section{Related Work}
\paragraph{Related Datasets}
End-to-end text-based QA has witnessed a surge in interest with the advent of large-scale datasets, which have been assembled based on \textsc{Freebase}~\cite{Berant13WebQuestions,Bordes2015SimpleQuestions}, \Wikipedia~\cite{Yang2015_WikiQA,Rajpurkar2016_SQUAD,hewlett2016_wikireading}, 
web search queries~\cite{Nguyen2016MSMARCO}, news articles~\cite{hermann2015teaching,Onishi2016Who}, books~\cite{Hill2015CBT,paperno2016lambada}, science exams~\cite{welbl2017crowdsourcing}, and trivia~\cite{BoydGraber_2012_Besting,dunn2017searchqa}.
Besides \emph{TriviaQA}~\cite{Joshi_2017_TriviaQA}, all these datasets are confined to single documents, and RC typically does not require a combination of multiple independent facts.
In contrast, \WikiHop and \MedHop are specifically designed for cross-document RC and multi-step inference.
There exist other multi-hop RC resources, but they are either very limited in size, such as the \emph{FraCaS} test suite, or based on synthetic language~\cite{Weston2015_bAbI}.
\TriviaQA partly involves multi-step reasoning, but the complexity largely stems from parsing compositional questions.
Our datasets center around compositional inference from comparatively simple queries and the cross-document setup ensures that multi-step inference goes beyond resolving co-reference.

\paragraph{Compositional Knowledge Base Inference}
Combining multiple facts is common for structured knowledge resources which formulate facts using first-order logic.
KB inference methods include Inductive Logic Programming~\cite{Quinlan1990_FOIL,Pazzani1991_FOCL,Richards1991_FORTE} and probabilistic relaxations to logic like Markov Logic~\cite{RichardsonDomingos2006_MLN,schoenmackers2008_scaling}.
These approaches suffer from limited coverage and inefficient inference, though efforts to circumvent sparsity have been undertaken~\cite{schoenmackers2008_scaling,Schoenmackers2010_WebHorn}.
A more scalable approach to composite rule learning is the Path Ranking Algorithm~\cite{lao2010PRA,lao2011random}, which performs random walks to identify salient paths between entities.
\newcite{Gardner2013EMNLP} circumvent these sparsity problems by introducing synthetic links via dense latent embeddings.
Several other methods have been proposed, using composition functions such as vector addition~\cite{Bordes2014_Subgraph}, RNNs~\cite{neelakantan2015compositional,das2016chains}, 
and memory networks~\cite{jain_2016_question}.
%
Another approach is the Neural Theorem Prover~\cite{rocktaschel2017NTP}, which uses dense rule and symbol embeddings to learn a differentiable  backward chaining algorithm.

All of these previous approaches center around learning how to combine facts from a KB, i.e.,\ in a structured form with pre-defined schema.
That is, they work as part of a pipeline, and either rely on the output of a previous IE step~\cite{Banko2008_OpenIE}, or on direct human annotation~\cite{Bollacker2008:FB} which tends to be costly and biased in coverage.
However, recent neural RC methods~\cite{Seo2016BidAF,Shen_2017_ReasoNet} have demonstrated that end-to-end language understanding approaches can infer answers directly from text -- sidestepping intermediate query parsing and IE steps.
Our work aims to evaluate whether end-to-end multi-step RC models can indeed operate on raw text documents only -- while performing the kind of inference most commonly associated with logical inference methods operating on structured knowledge.

\paragraph{Text-Based Multi-Step Reading Comprehension}
\newcite{Fried_2015_HigherOrder} have demonstrated that exploiting information from other related documents based on lexical semantic similarity is beneficial for re-ranking answers in open-domain non-factoid QA.
%
%
\newcite{Jansen2017Framing} chain textual background resources for science exam QA and provide multi-sentence answer explanations.
Beyond, a rich collection of neural models tailored towards multi-step RC has been developed.
Memory networks~\cite{weston_2014_memory,sukhbaatar_2015_N2N_MNN,Kumar2016_ICML} define a model class that iteratively attends over textual memory items, and they show promising performance on synthetic tasks requiring multi-step reasoning~\cite{Weston2015_bAbI}.
One common characteristic of neural multi-hop models is their rich structure that enables matching and interaction between question, context, answer candidates and combinations thereof~\cite{Peng2015_NN__Reasoning,Weissenborn_2016_Separating,Xiong_2016_Dynamic_coattention,Gated_memNN_Liu_2017}, which is often iterated over several times~\cite{Sordoni2016IterativeAN,neumann2016learning,Seo2017QRN,Hu_2917_Mnemonic} and may contain trainable stopping mechanisms~\cite{Graves_2016_Adaptive,Shen_2017_ReasoNet}.
All these methods show promise for single-document RC, and by design should be capable of integrating multiple facts across documents.
However, thus far they have not been evaluated for a cross-document multi-step RC task -- as in this work.

\paragraph{Learning Search Expansion}
Other research addresses expanding the document set available to a QA system, either in the form of web navigation~\cite{Nogueira_2016_WebNav}, or via query reformulation techniques, which often use neural reinforcement learning~\cite{Narasimhan_2016_Improving,Nogueira_2017_Query_Reformulation,Buck_2017_Ask}.
While related, this work ultimately aims at reformulating queries to better acquire evidence documents, and not at answering queries through combining facts.

\section{Conclusions and Future Work}
We have introduced a new cross-document multi-hop RC task, devised a generic dataset derivation strategy and applied it to two separate domains.
The resulting datasets test RC methods in their ability to perform composite reasoning -- something thus far limited to models operating on structured knowledge resources.
In our experiments we found that contemporary RC models can leverage cross-document information, but a sizeable gap to human performance remains.
Finally, we identified the selection of relevant document sets as the most promising direction for future research.
Thus far, our datasets center around factoid questions about entities, and as extractive RC datasets, it is assumed that the answer is mentioned verbatim.
While this limits the types of questions one can ask, these assumptions can facilitate both training and evaluation, 
and future work -- once free-form abstractive answer composition has advanced -- should move beyond. 
%
We hope that our work will foster research on cross-document information integration, working towards these long term goals.


\section*{Acknowledgments}
We would like to thank the reviewers and the action editor for their thoughtful and constructive suggestions, as well as Matko Bo\v{s}njak, Tim Dettmers, Pasquale Minervini, Jeff Mitchell, and Sebastian Ruder for several helpful comments and feedback on drafts of this paper.
This work was supported by an Allen Distinguished Investigator Award, a Marie Curie Career Integration Award, the EU H2020 SUMMA project (grant agreement number 688139), and an Engineering and Physical Sciences Research Council scholarship.

\bibliography{main.bib}

\bibliographystyle{acl2012}

\clearpage

\appendix

\section{Appendix: Versions}
This paper directly corresponds to the TACL version,\footnote{\url{https://transacl.org/ojs/index.php/tacl/article/view/1325}} apart from minor changes in wording, additional footnotes, and these appendices.

\section{Appendix: Candidate and Document statistics}\label{appendix:plots}

\begin{figure}[t]
    \centering
    \includegraphics[width=0.9\columnwidth]{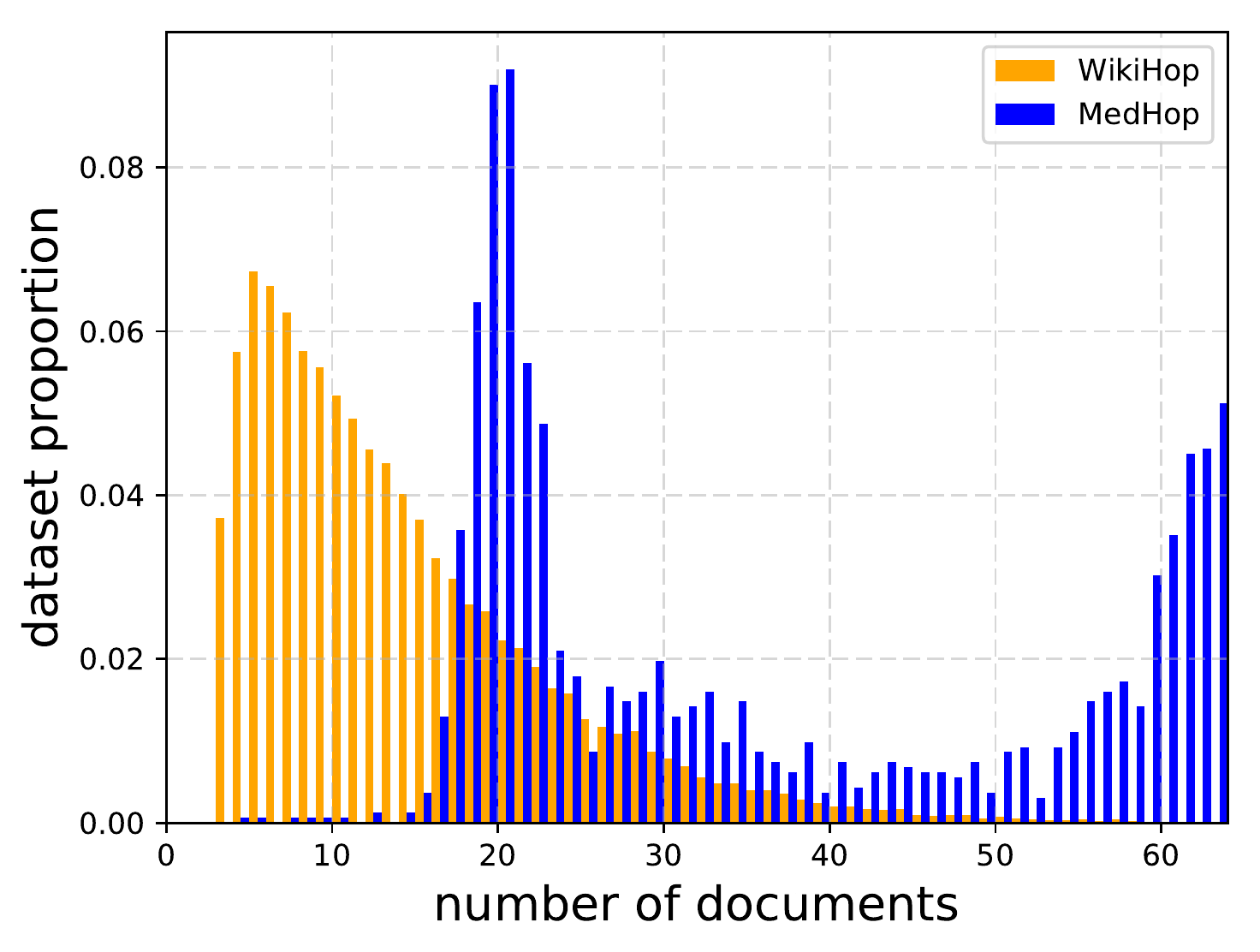}
    \caption{Support documents per training sample.} \label{fig:num_supports}
\end{figure}

Figure~\ref{fig:num_supports} illustrates the distribution of the number of support documents per sample.
\WikiHop shows a Poisson-like behaviour -- most likely due to structural regularities in \Wikipedia -- whereas \MedHop exhibits a bimodal distribution, in line with our observation that certain drugs and proteins have far more interactions and studies associated with them.

\begin{figure}[t]
     \centering
     \includegraphics[width=0.9\columnwidth]{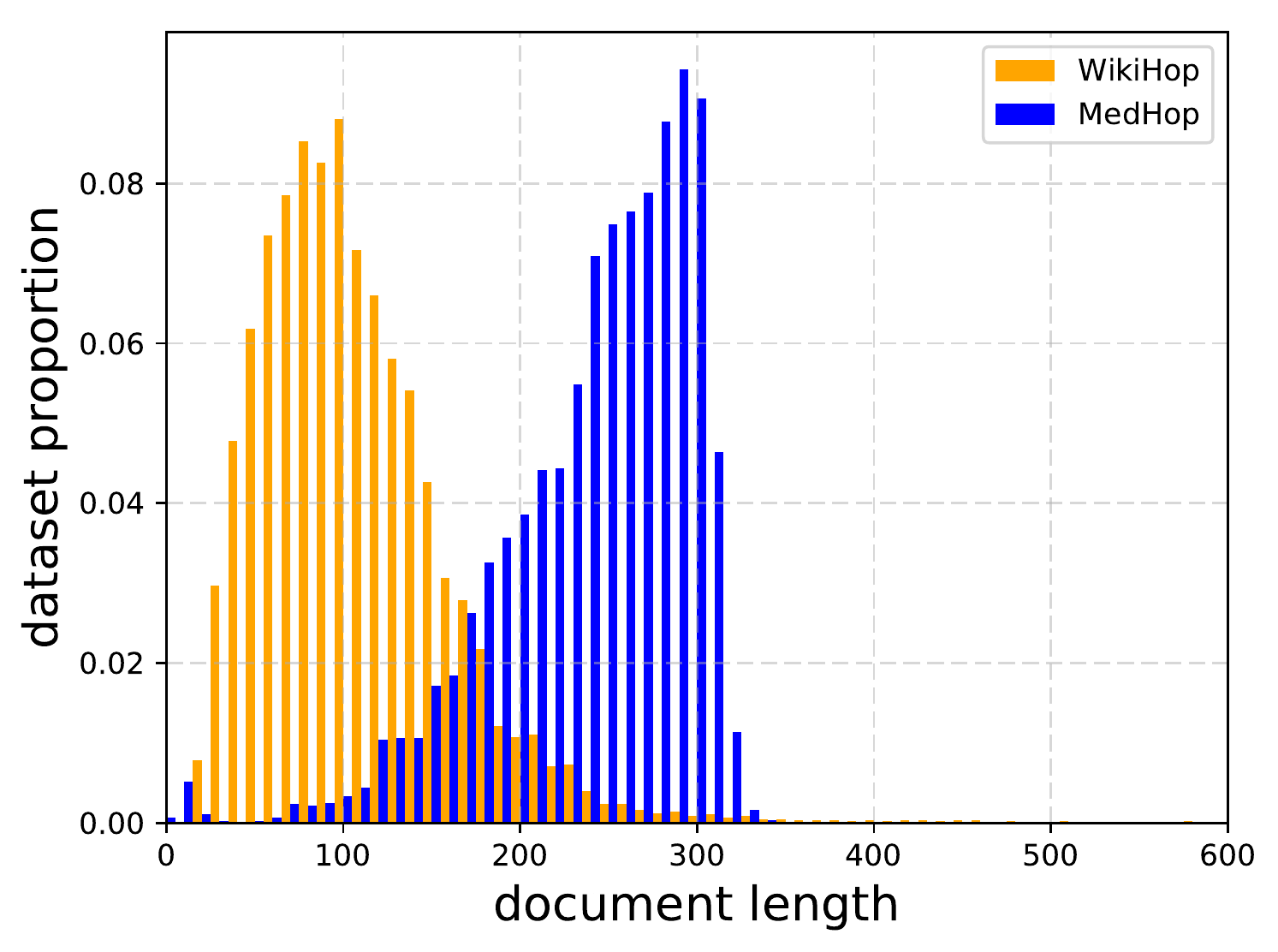}
     \caption{
         Histogram for document lengths in \WikiHop and \MedHop.
     }
     \label{fig:document_lengths}
 \end{figure}

Figure~\ref{fig:document_lengths} shows the distribution of document lengths for both datasets.
Note that the document lengths in \WikiHop correspond to the lengths of the first paragraphs of \Wikipedia articles.
\MedHop on the other hand reflects the length of research paper abstracts, which are generally longer.

\begin{figure}[t]
    \centering
    \includegraphics[width=0.9\columnwidth]{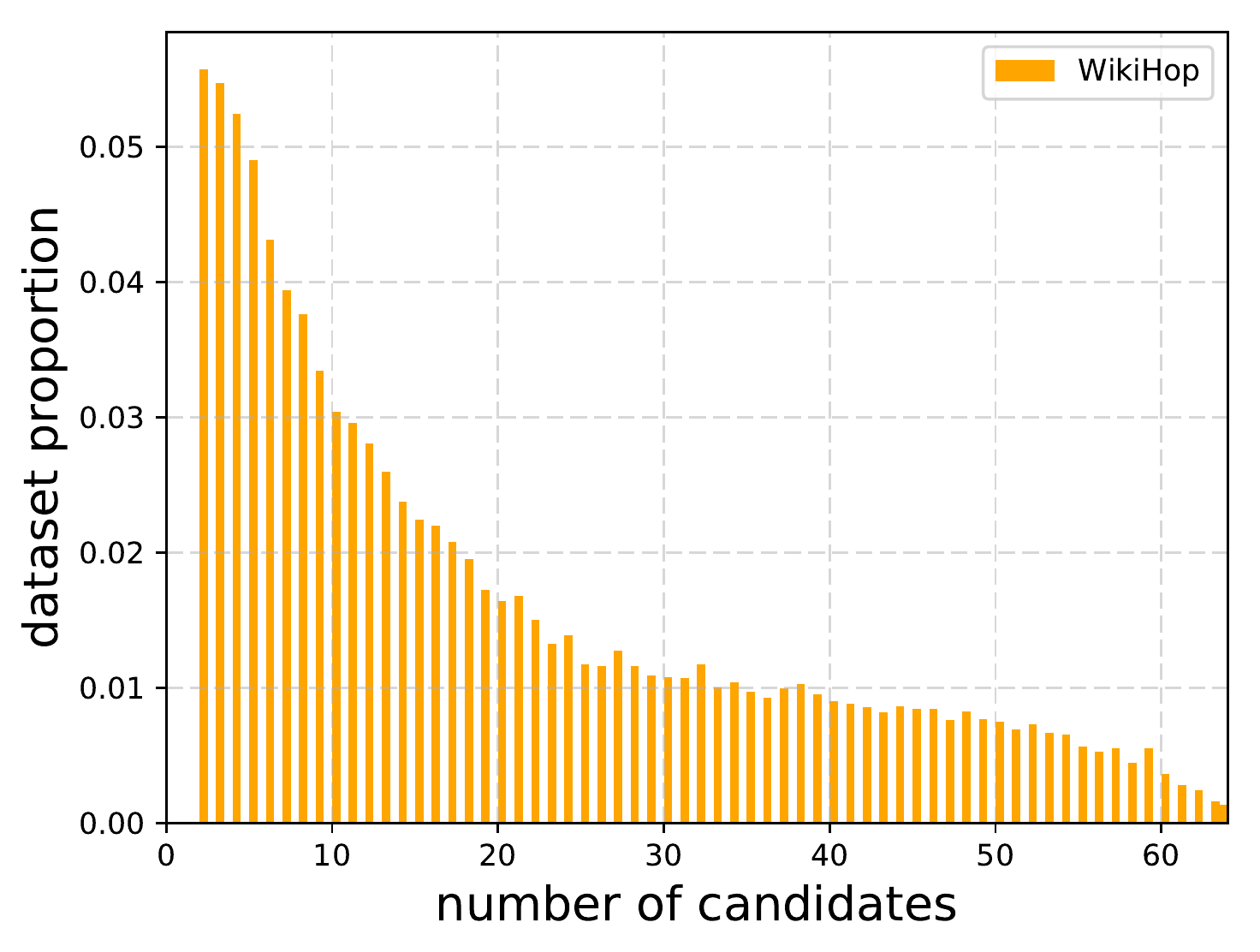}
    \caption{
        Histogram for the number of candidates per sample in \WikiHop.
    }
    \label{fig:candidate_histogram}
\end{figure}

Figure~\ref{fig:candidate_histogram} shows a histogram with the number of candidates per sample in \WikiHop, and the distribution shows a slow but steady decrease.
For \MedHop, the vast majority of samples have 9 candidates, which is due to the way documents are selected up until a maximum of 64 documents is reached.
Very few samples have fewer than 9 candidates, and samples would have far more false candidates if more than 64 support documents were included.
%

\section{Appendix: Document-Cue examples}\label{appendix:cue-examples}
\begin{table*}
    \centering
    \begin{tabular}{p{.15\textwidth} p{.7\textwidth} p{.05\textwidth} p{.05\textwidth}}
        \textbf{Answer} $a$      & \textbf{Wikipedia article} $d$ & \textbf{Count}  & \textbf{Prop.} \\
        \toprule
        united states of america    & A \textbf{U.S. state} is a constituent political entity of the United States of America. & 68,233 & 12.9\%   \\
        \midrule
        united kingdom    & \textbf{England} is a country that is part of the United Kingdom. & 54,005 & 10.2\%   \\
        \midrule
        taxon    & In biology, a \textbf{species} (abbreviated sp., with the plural form species abbreviated spp.) is the basic unit of biological classification and a taxonomic rank. & 40,141  & 7.6\%    \\
        \midrule
        taxon    &  A \textbf{genus} (pl. \textbf{genera}) is a taxonomic rank used in the biological classification & 38,466 & 7.3\%    \\
        \midrule
        united kingdom    & The \textbf{United Kingdom of Great Britain and Northern Ireland}, commonly known as the \textbf{United Kingdom (UK)} or Britain, is a sovereign country in western Europe.  & 31,071  & 5.9\%  \\
        \midrule
        taxon    & \textbf{Biology} is a natural science concerned with the study of life and living organisms, including their structure, function, growth, evolution, distribution, identification and taxonomy. & 27,609   & 5.2\%  \\
        \midrule
        united kingdom    &  \textbf{Scotland} [...] is a country that is part of the United Kingdom and covers the northern third of the island of Great Britain.  & 25,456  & 4.8\%     \\
        \midrule
        united kingdom    & \textbf{Wales} [...] is a country that is part of the United Kingdom and the island of Great Britain.  & 21,961 & 4.2\%     \\
        \midrule
        united kingdom    & \textbf{London} [...] is the capital and most populous city of England and the United Kingdom, as well as the most populous city proper in the European Union. & 21,920  & 4.2\%  \\
        \midrule
        ... &... &... \\
        \midrule
        united states of america    &  \textbf{Nevada} (Spanish for "snowy"; see pronunciations) is a state in the Western, Mountain West, and Southwestern regions of the United States of America. & 18,215  & 3.4\%  \\
        \midrule
        ... &... &... \\
        \midrule        
        italy    &  The \textbf{comune} [...] is a basic administrative division in Italy, roughly equivalent to a township or municipality.  & 8,785 &  1.7\%  \\ 
        \midrule        
        ... &... &... \\
        \midrule        
        human settlement     &  A \textbf{town} is a human settlement larger than a village but smaller than a city.   & 5,092 & 1.0\%   \\ 
        \midrule        
        ... &... &... \\
        \midrule  
        people's republic of china & \textbf{Shanghai} [...] often abbreviated as Hu or Shen, is one of the four direct-controlled municipalities of the People's Republic of China. & 3,628 & 0.7\% \\ 
        \bottomrule
    \end{tabular}
    \caption{
      Examples with largest $\text{\emph{cooccurrence}}(d,c)$ statistic, before filtering.
      The \emph{Count} column states $\text{\emph{cooccurrence}}(d,c)$; the last column states the corresponding relative proportion of training samples (total 527,773).
    }
    \label{tbl:doc-cue-examples}
\end{table*}
Table~\ref{tbl:doc-cue-examples} shows examples of answers and articles which, before filtering, frequently appear together in \WikiHop.

\section{Appendix: Gold Chain Examples}
Table~\ref{tbl:gold-chain-examples} shows examples of document gold chains in \WikiHop.
Note that their lengths differ, with a maximum of 3 documents.

\section{Appendix: Query Types}
Table~\ref{tbl:query-types} gives an overview over the 25 most frequent query types in \WikiHop and their relative proportion in the dataset.
Overall, the distribution across the 277 query types follows a power law.
%


\begin{table*}
    \centering
    \begin{tabular}{p{\textwidth}}
        \toprule
        \textbf{Query:} (the big broadcast of 1937, genre, ?)\\
        \textbf{Answer:} musical film\\
        \textbf{Text 1:} The \textbf{Big Broadcast of 1937} is a 1936 Paramount Pictures production directed by Mitchell Leisen, and is the third in the series of Big Broadcast movies. 
        The musical comedy stars Jack Benny, George Burns, Gracie Allen, Bob Burns, Martha Raye, Shirley Ross [...]\\
        \textbf{Text 2:} \textbf{Shirley Ross} (January 7, 1913 – March 9, 1975) was an American actress and singer, notable for her duet with Bob Hope, "Thanks for the Memory" from "The Big Broadcast of 1938"[...]\\
        \textbf{Text 3:} \textbf{The Big Broadcast of 1938} is a Paramount Pictures \underline{musical film} featuring W.C. Fields and Bob Hope. 
        Directed by Mitchell Leisen, the film is the last in a series of "Big Broadcast" movies[...]\\
        \midrule
        \textbf{Query:} (cmos, subclass\_of, ?)\\ 
        \textbf{Answer:} semiconductor device \\
        \textbf{Text 1:} \textbf{Complementary metal-oxide-semiconductor} (CMOS) [...] is a technology for constructing integrated circuits. [...] CMOS uses complementary and symmetrical pairs of p-type and n-type metal oxide semiconductor field effect transistors (MOSFETs) for logic functions. [...]\\
        \textbf{Text 2:} A \textbf{transistor} is a \underline{semiconductor device} used to amplify or switch electronic signals[...]\\
        \midrule
        \textbf{Query:} (raik dittrich, sport, ?) \\ 
        \textbf{Answer:} biathlon \\
        \textbf{Text 1:} Raik Dittrich (born October 12, 1968 in Sebnitz) is a retired East German biathlete who won two World Championships medals. He represented the sports club SG Dynamo Zinnwald [...]\\
        \textbf{Text 2:} SG Dynamo Zinnwald is a sector of SV Dynamo located in Altenberg, Saxony[...] The main sports covered by the club are \underline{biathlon}, bobsleigh, luge, mountain biking, and Skeleton (sport)[...] \\
        \midrule
        \textbf{Query:} (minnesota gubernatorial election, office\_contested, ?) \\ 
        \textbf{Answer:} governor \\
        \textbf{Text 1:}
        The 1936 Minnesota gubernatorial election took place on November 3, 1936. Farmer-Labor Party candidate Elmer Austin Benson defeated Republican Party of Minnesota challenger Martin A. Nelson. \\
        \textbf{Text 2:}
        \textbf{Elmer Austin Benson} [...] served as the 24th \underline{governor} of Minnesota, defeating Republican Martin Nelson in a landslide victory in Minnesota's 1936 gubernatorial election.[...]\\
        \midrule
        \textbf{Query:} (ieee transactions on information theory, publisher, ?) \\ 
        \textbf{Answer:} institute of electrical and electronics engineers\\
        \textbf{Text 1:} \textbf{IEEE Transactions on Information Theory} is a monthly peer-reviewed scientific journal published by the IEEE Information Theory Society [...] the journal allows the posting of preprints [...]\\
        \textbf{Text 2:} 
        The \textbf{IEEE Information Theory Society} (ITS or ITSoc), formerly the IEEE Information Theory Group, is a professional society of the \underline{Institute of Electrical and Electronics Engineers (IEEE)} [...]\\
        \midrule
        \textbf{Query:} (country\_of\_citizenship, louis-philippe fiset, ?)\\
        \textbf{Answer:} canada\\
        \textbf{Text1:} \textbf{Louis-Philippe Fiset} [...] was a local physician and politician in the Mauricie area [...] \\
        \textbf{Text2:} Mauricie is a traditional and current administrative region of Quebec. La Mauricie National Park is contained within the region, making it a prime tourist location. [...] \\
        \textbf{Text3:} La Mauricie National Park is located near Shawinigan in the Laurentian mountains, in the Mauricie region of Quebec, \underline{Canada} [...]\\
        \bottomrule
    \end{tabular}
    \caption{Examples of document gold chains in \WikiHop. Article titles are boldfaced, the correct answer is underlined.
    }
    \label{tbl:gold-chain-examples}
\end{table*}


\begin{table*}\label{tab:query_types}
    \centering
    \begin{tabular}{l r}
        Query Type  & Proportion in Dataset\\
        \toprule
        instance\_of & 10.71 \% \\
        located\_in\_the\_administrative\_territorial\_entity & 9.50 \% \\
        occupation & 7.28 \% \\
        place\_of\_birth & 5.75 \% \\
        record\_label & 5.27 \% \\
        genre & 5.03 \% \\
        country\_of\_citizenship & 3.45 \% \\
        parent\_taxon & 3.16 \% \\
        place\_of\_death & 2.46 \% \\
        inception & 2.20 \% \\
        date\_of\_birth & 1.84 \% \\
        country & 1.70 \% \\
        headquarters\_location & 1.52 \% \\
        part\_of & 1.43 \% \\
        subclass\_of & 1.40 \% \\
        sport & 1.36 \% \\
        member\_of\_political\_party & 1.29 \% \\
        publisher & 1.16 \% \\
        publication\_date & 1.06 \% \\
        country\_of\_origin & 0.92 \% \\
        languages\_spoken\_or\_written & 0.92 \% \\
        date\_of\_death & 0.90 \% \\
        original\_language\_of\_work & 0.85 \% \\
        followed\_by & 0.82 \% \\
        position\_held & 0.79 \% \\
        \midrule
        Top 25 & 72.77 \% \\
        Top 50 & 86.42 \% \\
        Top 100 & 96.62 \% \\
        Top 200 & 99.71 \% \\
        \bottomrule
    \end{tabular}
    \caption{The 25 most frequent query types in \WikiHop alongside their proportion in the training set.
    }
    \label{tbl:query-types}
\end{table*}

\end{document}